%% file: main.tex
\documentclass[runningheads]{llncs}

\usepackage{eccv}

\usepackage{eccvabbrv}

\usepackage{graphicx}
\usepackage{booktabs}
\usepackage{url}
\usepackage{amssymb}
\usepackage{pifont}
\usepackage{bbding}
\usepackage{xcolor}
\usepackage{amsmath}
\usepackage{float}
\usepackage{placeins}
\usepackage{colortbl}
\usepackage{courier}
\usepackage{caption}
\usepackage{multirow}
\usepackage{wrapfig}
\usepackage[ruled,linesnumbered,noend,vlined]{algorithm2e}

\newcommand{\cmark}{\ding{51}}
\newcommand{\xmark}{\ding{55}}
\newcommand{\model}{MemTree3D}
\newcommand{\boldmodel}{\textbf{\model}}

\usepackage[accsupp]{axessibility}  % Improves PDF readability for those with disabilities.

\usepackage[pagebackref,breaklinks,colorlinks,citecolor=eccvblue]{hyperref}
\usepackage{orcidlink}

\begin{document}

\title{Memory Tree Guided Key Frame Querying for Efficient 3D Question Answering} 

\titlerunning{Memory Tree 3D}

% TODO FINAL: Replace ORCID placeholders with the authors' ORCID IDs.
% \author{Hsiang-Wei Huang\inst{1,2}\thanks{Work done during internship at Amazon.}\orcidlink{0009-0009-2474-8869} \and
% Fu-Chen Chen\inst{2}\orcidlink{0000-0002-6396-2798} \and
% Li-Wu Tsao\inst{2}\orcidlink{0009-0009-0935-3842} \and
% Cheng-Han Lee\inst{3}\orcidlink{0009-0008-3260-258X} \and
% Che-Chun Su\inst{2}\orcidlink{0009-0008-4555-2032} \and
% Lu Xia\inst{2}\orcidlink{0000-0000-0000-0000} \and
% Ronghui Peng\inst{2}\orcidlink{0000-0002-7648-2988} \and
% Jenq-Neng Hwang\inst{1}\orcidlink{0000-0002-8877-2421} \and
% Min Sun\inst{2}\orcidlink{0000-0000-0000-0000} \and
% Cheng-Hao Kuo\inst{2}\orcidlink{0000-0000-0000-0000}}

\author{Hsiang-Wei Huang\inst{1,2}\thanks{Work done during internship at Amazon.} \and
Fu-Chen Chen\inst{2} \and
Li-Wu Tsao\inst{2} \and
Cheng-Han Lee\inst{3} \and
Che-Chun Su\inst{2} \and
Lu Xia\inst{2} \and
Ronghui Peng\inst{2} \and
Jenq-Neng Hwang\inst{1} \and
Min Sun\inst{2} \and
Cheng-Hao Kuo\inst{2}}

\authorrunning{H.-W.~Huang et al.}

\institute{University of Washington, United States \and
Amazon, United States \and
The University of Texas at Austin, United States}

\maketitle
\input{sec/0_abstract}    
\input{sec/1_intro}
\input{sec/2_survey}
\input{sec/3_method}
\input{sec/4_experiments}
\input{sec/6_conclusion}


\clearpage

\bibliographystyle{splncs04}
\bibliography{main}

\input{sec/X_suppl}

\end{document}

%% file: sec/0_abstract.tex
\begin{abstract}
Answering questions accurately and efficiently in embodied scenarios presents significant challenges due to limited computational and memory resources for Vision Language Model~(VLM) inference. Existing methods adopt visual search key frame retrieval method to select critical question-related key frames for VLM input. However, visual search methods are inefficient because they require visual search among thousands of video frames for each individual user query. In this work, we propose a memory tree guided key frame selection paradigm for efficient 3D question answering in embodied scenarios. Our method leverages a compact and reusable 3D scene representation, termed \model, which supports real-time online construction leveraging camera 6-DoF poses. \model~captures multi-level 3D scene information, enabling a Large Language Model to efficiently query and retrieve question-relevant key frames through our scoring-based frame selection without reprocessing the entire video stream. On OpenEQA, our method improves the LLM-Match of GPT-4o by 17.4\%, LLaVA-OneVision-7B by 5.8\%, outperforms existing visual search methods. Our code is available at \href{https://github.com/hsiangwei0903/MemTree3D}{https://github.com/hsiangwei0903/MemTree3D}.

\keywords{3D Question Answering \and Embodied Question Answering}

\end{abstract}

%% file: sec/1_intro.tex
\section{Introduction}
3D question answering~\cite{openeqa,scanqa,sqa3d} is a fundamental task in the field of scene understanding. It plays a crucial role in bridging vision–language model systems with real-world 3D environments, enabling a wide range of downstream applications such as embodied navigation~\cite{3dmem,openeqa,chen2020soundspaces}, household robotics~\cite{soni2024advancing}, augmented and virtual reality~\cite{lin2025samr,duan2025advancing}, and assistive agents~\cite{das2018embodied,shridhar2020alfred,huang2025warehouse} that require accurate spatial reasoning and grounded semantic understanding of complex scenes. However, performing 3D question answering efficiently and accurately has been challenging for real-world embodied systems due to restricted computational and GPU memory resources. Current state-of-the-art Vision-Language Models (VLMs) mostly adopt the typical transformer architecture~\cite{vaswani2017attention}, with the attention mechanism's computation and memory requirements scaling quadratically with the input video length. This limitation creates a fundamental bottleneck, as existing 3D scene scan videos are typically captured at high FPS, which results in thousands of video frames and necessitates the use of frame sampling strategies for VLM-based 3D question answering.

\input{fig/teaser}

Existing LLM/VLM-based 3D question answering methods can be broadly categorized into several mainstream approaches, including (a) Multi-Frame VLMs, (b) Scene-Graph Method, and (c) Visual Search Method, as shown in Figure~\ref{fig:teaser}. State-of-the-art multi-frame VLMs~\cite{llava-onevision,llava-next-interleave} already possess the ability to perform 3D question answering, but are largely constrained by the computational and memory resources in the embodied scenario, often require video temporal downsampling strategies that leads to visual information loss. Scene-graph methods~\cite{scenegpt,gu2024conceptgraphs,openeqa} leverage object-level scene representation for LLM question answering, the object-only representation, despite being simple and memory efficient, limits the system from performing fine-grained visual recognition beyond object-level understanding. To achieve a balance between memory constraint and more advanced visual recognition ability, visual search methods~\cite{xu2024vlmgrounder} adopt an LLM to first perform query analysis, which identifies query-related objects and performs key object visual search in all video frames to select key frames for VLM input. This strategy reduces the memory overhead of VLM while preserving key frame information, yet still possess several major limitations, including 1) the visual search runtime scales with the video length, which hinders its ability for real-time user interaction 2) the visual search is required for each individual query, and cannot efficiently handle multi-round query scenarios, 3) the method does not account for visual search failure, where the vision model fails to retrieve meaningful query-related key frames from the video.

In this work, we identify and address two major limitations of existing visual search–based 3D question answering methods. First, these approaches rely heavily on the underlying vision model to perform visual search, yet lack a mechanism to recover from failures when query-relevant objects are missed. Second, they require exhaustive visual inference over all video frames, which must be repeated for every input query and significantly hinders real-time response. To alleviate these issues, we introduce an LLM-driven key-frame retrieval framework built upon a pre-constructed 3D scene representation. This design enables more flexible and robust query handling through the LLM’s explicit spatial and temporal reasoning capabilities. Furthermore, the system efficiently retrieves relevant key frames with respect to the user query without extensive visual search. Specifically, we leverage a tree-based architecture as a compact, reusable representation of the 3D scene, which can be efficiently queried by the LLM to generate informative cues for retrieving query-relevant key frames. These selected frames are then passed to a VLM to perform final question answering.

Our approach begins with the online and real-time construction of a hierarchical representation \boldmodel. \model~encodes 3D scene at multiple levels, including frame-level detections, temporally-aware object relationships, and spatially localized segments derived from 6-DoF camera poses. This tree-based structure enables the LLM to reason over the scene content and identify critical frames without scanning every video frame. It is also reusable across multi-round queries: once constructed, the tree can be queried repeatedly by the LLM for different questions without re-running the vision model. Finally, it is robust to visual search failures. Because the LLM reasons over the symbolic and structural information in the tree~(an example in Fig~\ref{fig:query}), it can still identify relevant frames even when query-related objects are missing. Extensive experiments on 3D question answering benchmarks show that our MemTree3D key frame selection paradigm improves state-of-the-art question answering accuracy, outperforms uniform sampling GPT-4o by 17.4\%, largely reduces the runtime compared to visual search method by efficiency through \model~querying and enhances robustness to perception failures via LLM-based reasoning. 

In summary, the contribution of the work are three-folds:

\begin{enumerate}
\item We introduce \boldmodel, a compact and reusable 3D scene representation that supports real-time construction and enables LLM-driven key-frame selection, paving the way for scalable real-world embodied applications.
\item We propose a novel \model~guided key frame selection paradigm for the 3D question answering task, significantly improving efficiency over existing visual search key-frame selection approaches.
\item We conduct extensive experiments and show that our method improves the LLM-Match of GPT-4o by 17.4\% and achieves a 69.2\% key-frame retrieval speedup compared to prior visual search methods, demonstrating a strong and efficient paradigm for 3D question answering.
\end{enumerate}

%% file: fig/teaser.tex
\begin{figure}[t]
    \centering
    \includegraphics[width=0.98\linewidth]{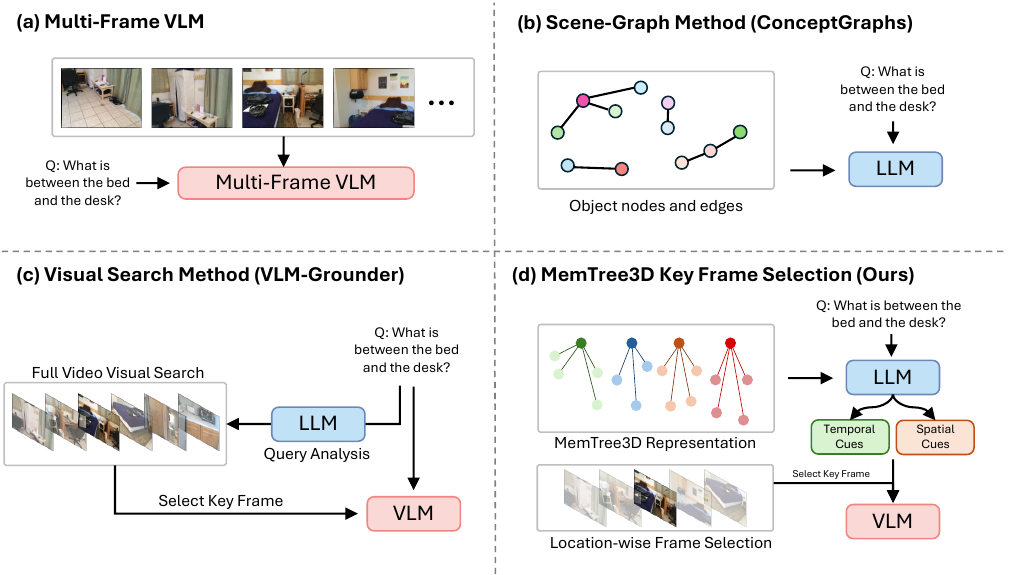}
    \caption{\textbf{Comparison of 3D question answering methods.} (a) Multi-Frame VLM takes 3D scan video as input but suffers from proportional computation and memory cost scales with video length. (b) Scene-Graph Methods such as ConceptGraphs~\cite{gu2024conceptgraphs} leverages object-level scene graph and an LLM for question answering, but the coarse-level scene representation leads to visual information loss. (c) Visual Search Method such as VLM-Grounder~\cite{xu2024vlmgrounder} leverages object detector to perform visual search for each query, which leads to extensive runtime. (d) Our method adopts an LLM to reason over our designed MemTree3D representation and provide temporal and spatial cues for efficient, location-wise frame selection, bypassing the extensive visual search.}
    \label{fig:teaser}
\end{figure}

%% file: sec/2_survey.tex
\section{Related Work}
\subsection{3D Question Answering}
Capturing complex 3D scenes in a compact representation and conduct question answering remains a significant challenge in current research. Recent 3D-LLMs~\cite{3d-vista,zhu2024unifying,xu2024pointllm} leverage point cloud representations as LLM input for question answering task, but they require extensive 3D data training and still fall short in performance and generalizability compared to 2D models~\cite{dtc,wang2025d,zhang2024agent3d}. Object-centric representations~\cite{chat-3d,chatscene,multiply} and 3D scene graphs~\cite{gu2024conceptgraphs,3dscenegraph,cherian20222,Wu_2021_CVPR,scenegpt} are also popular approaches, yet they suffer from perception failures such as missing objects or false positives. The coarse, object-level representation also limits scene-graph's ability in answering questions related to 3D scene details. Furthermore, most existing scene graphs have inherently complex construction process, which require multiple foundation models~\cite{sam,groundingdino,clip} and often cannot support real-time construction. On the other hand, recent 2D video VLM~\cite{zheng2025video,llava3d,videochat,videochat2,videollama,videollama2} directly utilize 3D scan videos as input for 3D scene question answering, the extensive training data enables 2D VLMs to achieve comparable performance with 3D-LLMs. But they face significant computational overhead due to the excessive number of frames in high FPS 3D scan videos. To address these challenges, we propose \model, a lightweight 3D scene representation that differs from traditional scene graphs by incorporating temporal information that supports key frame retrieval and real-time 25+ FPS construction. Rather than serving as the final scene representation, our proposed \model~acts as an intermediate structure for LLM-driven key frame retrieval, mitigating the visual information loss in existing scene graph-based methods, and further address the challenges of computational overhead in video-LLMs by selecting question-related key frame for 3D question answering.

\subsection{Key Frame Selection for Video Understanding}
To address the computational and memory overhead in long-form video, recent research has developed key frame selection techniques to select question-relevant frames for VLM input. Most existing key frame selection methods follow the visual search key frame selection paradigm, which leverages the input question to identify relevant frames using techniques such as detecting question-related objects~\cite{tstar,xu2024vlmgrounder}, VLM-based selection~\cite{hu2025cos,3dmem}, or computing image-text similarity scores~\cite{wang2025d} via vision foundation models~\cite{clip,blip2}. While these approaches generally outperform naive uniform sampling, the visual search paradigm significantly limits their efficiency in multi-round user query scenarios, as each query requires re-running visual search across the entire video. In this work, we investigate the setting of leveraging a one-time, pre-constructed high-level 3D scene representation for efficient, LLM-driven key frame retrieval. Our approach deviates from standard visual search key frame selection paradigm that suffers from the large computational cost scaling with the video length, our paradigm achieves highly efficient key frame selection that does not significantly degrade with the video length, while achieving SOTA 3D question answering performance.

% \paragraph{3D Question Answering MLLM.}
% Existing 3D question answering MLLMs can be broadly categorized into video-LLM and 3D-LLM, with the former taking the 3D scan video as input, while the latter consumes 3D scene point cloud or voxel representation. Video-LLMs~\cite{videochat,videochatgpt,videollama,llavanext} can be directly applied to the 3D question answering task under zero-shot setting, yet they suffer when handling a large number of frames, which incur extensive memory and computational cost. 3D-LLM methods~\cite{scenellm,3dllm,leo,3d-vista,ll3da,xu2024pointllm}, despite using a more compact point cloud or voxel representations, require extensive fine-tuning on 3D data, and do not achieve significant performance advantages when compared with zero-shot, 2D video-LLM~\cite{dtc,wang2025d}.

%% file: sec/3_method.tex
\section{Method}

\subsection{Overview}
We propose the \model~guided key-frame selection paradigm for the 3D question answering task, motivated by the need to bypass the extensive visual inference required by existing visual search–based methods, which must run a vision model over all video frames for every individual query. We illustrate the overview of our method in Fig~\ref{fig:query}. Given a 3D scan video, we first pre-construct a compact and lightweight 3D scene representation \model~from the video. An LLM then takes the query together with \model~as input to perform explicit reasoning and generate spatial and temporal cues for efficient key-frame retrieval. The selected key frames are subsequently passed to a VLM for final question answering. This paradigm offers two key advantages. First, it reduces overreliance on visual search results by enabling flexible, query-adaptive cue generation through LLM reasoning. Furthermore, it improves efficiency by eliminating repeated, query-dependent visual inference that scales with video length, replacing it with an LLM-based cue generation step with consistent runtime.

\subsection{\model}
\label{sec:construction}

\input{fig/query}

\input{fig/method}

\input{alg/location}

We first introduce our proposed \model~representation, a compact and lightweight 3D scene representation designed to support LLM reasoning and to provide spatial and temporal cues for key-frame selection. Despite the existence of several well-developed 3D scene representations, including point clouds~\cite{3dllm,xu2024pointllm}, feature fields~\cite{multiply,scenellm}, and scene graphs~\cite{conceptfusion,gu2024conceptgraphs}, each exhibits limitations that make it unsuitable for our method. Point cloud and feature field representations typically require additional LLM retraining, which incurs substantial computational and development cost. Existing scene graph–based methods such as ConceptGraphs~\cite{gu2024conceptgraphs}, while capable of zero-shot LLM reasoning through textual serialization, rely on an extremely heavyweight construction pipeline that often combines multiple foundation models~\cite{sam,clip,groundingdino} and fail to achieve 10+ FPS real-time performance. Motivated by these limitations, we design \boldmodel, a \textbf{minimal} and \textbf{lightweight} 3D scene representation that supports real-time~(\textbf{25+ FPS}) construction while remaining \textbf{sufficient} for effective LLM reasoning and cue generation, which offers practicability for real-world deployment. \model~is organized as a three-level tree structure. We describe the design and functionality of each level as follows:

% \vspace{4pt}

\noindent{\textbf{Location Node.}}
We split the 3D scene video into multiple segments and construct a location node~(\texttt{LocNode}) for each segment, as shown in Fig~\ref{fig:method}. Each \texttt{LocNode} is assigned a unique location ID. While several existing works~\cite{he2024ma,wang2025videotree} rely on image semantic to cluster video into segments, such approaches pose challenges in the embodied AI setting. These clustering methods are effective in general video question answering tasks, where inter-frame differences often correspond to changes in video content or camera shots. However, in the embodied scenario, the camera captures a continuous 3D scan with smaller semantic variation between adjacent frames, especially in visually uniform scene. This makes semantic-based clustering less reliable for identifying spatial transitions in 3D scenes. The second challenge lies in the computational overhead of extracting visual semantics, which relies on off-the-shelf vision foundation model. This further poses difficulties in resource-constrained embodied scenarios.

To address these limitations, we propose to segment the video stream in a way that (1) introduces minimal additional computational cost, and (2) ensures that each segment corresponds to a distinct location in the 3D scene. In our work, we utilize the 6-DoF pose, which is information that can be obtained directly from most existing embodied AI devices. We keep track of each frame's 6-DoF pose during video processing and maintain a previous 6-DoF pose, $P_{prev}$, to calculate the translation and rotation with respect to the 6-DoF pose at the current frame, $P_t$. Whenever the 6-DoF changes exceed translation threshold $T_{thres}$ or rotation threshold $R_{thres}$, we construct a new \texttt{LocNode} and update $P_{prev}$ with the current pose $P_t$; we provide a detailed pseudo code in Algorithm.~\ref{alg:location}.\\

\noindent{\textbf{Object Node.}} In the 3D scene scan, we continuously leverage an object detector~\cite{yoloworld} and a multi-object tracker~\cite{aharon2022bot} to obtain object detections and their temporal dependencies across time stamps. In each location node, we can obtain multiple object tracklets from the detector and tracker. For each location, we wrapped each observed tracklet into an object node \texttt{ObjNode}. Each \texttt{ObjNode} stores a compact trajectory derived from the tracker. The combinations of \texttt{ObjNode} at each location ensure a compact, high-level coarse representation, which can enable the LLM querying to reason over the coarse perception content for each location in the scene.\\

\noindent{\textbf{Detection Node.}}
Detection node~(\texttt{DetNode}) is the leaf component in the \model~structure, representing the detection in each frame. Each \texttt{DetNode} contains only the basic attribute, including the detection time-stamp, detection results like bounding box coordinates, and detection confidence representing the quality of the detection, which will be used in the later frame selection.

\subsection{LLM Spatial and Temporal Cue Generation}
\label{sec:llm query}
After constructing \model, we obtain a compact 3D scene representation suitable for LLM reasoning. Given a user question $\textit{Q}$, we leverage an LLM to reason over \model~and generate temporal and spatial cues for subsequent key-frame selection. Specifically, we serialize the constructed \model~representation into a textual JSON format and feed it to the LLM, as illustrated in Fig~\ref{fig:query}. This JSON representation includes only the first two levels of nodes, namely \texttt{LocNode} and \texttt{ObjNode}, while \texttt{DetNode} is omitted since fine-grained detection information is not essential for high-level cue generation. The LLM performs explicit reasoning based on the objects present in each location and to produce temporal and spatial cues that guide efficient frame retrieval.\\

\noindent{\textbf{Temporal Cues.}}
The temporal cue generation focuses on identifying a small set of candidate locations (and their associated temporal segments) that are most likely to contain the answer to the query. When query-relevant objects are present, the LLM can directly associate the question with the corresponding locations by matching semantic object attributes. In more challenging cases where critical objects are missing due to perception failures, the LLM reasons over the relational and contextual information encoded in \model~to infer the most plausible locations that may contain the answer. This reasoning-driven temporal cue generation enables robust recovery from perception errors, a capability that differentiates our work with existing visual search~\cite{xu2024vlmgrounder} or scene-graph–based approaches~\cite{conceptfusion,gu2024conceptgraphs}, which typically fail when the underlying vision model cannot identify key objects.\\

\noindent{\textbf{Spatial Cues.}}
In addition to temporal cues, the LLM generates spatial cues by predicting two sets of objects: key objects $O_{\text{key}}$ and cue objects $O_{\text{cue}}$. Key objects correspond to entities that are directly relevant for locating the answer to the question, while cue objects denote surrounding or co-occurring objects that are likely to appear near the key objects in the scene. These spatial cues capture both direct and contextual object relationships and are subsequently used as criteria in our scoring-based key-frame selection stage. By explicitly modeling spatial relationships through LLM reasoning, our approach reduces overreliance on exact object detections and provides flexible, query-adaptive guidance for efficient frame retrieval.

% In the scenario where the object in the user question is detected, LLM can directly predict the correct locations without extensive reasoning. However, when an object is not detected due to a detection failure, our framework demonstrates its robustness through the LLM reasoning over the \model~and predicts the most possible locations that contain the answer to the query. This distinct our work from existing visual search and scene-graph methods, which are not able to answer the question correctly when the vision model fails to identify the critical objects. Besides predicting the top $k$ possible locations, we also prompt the LLM to provide two object lists, including key objects $O_{key}$ and cue objects $O_{cue}$. Key objects $O_{key}$ are objects that can help locate the answer to the question, and cue objects $O_{cue}$ are objects that might be close or near the key objects, these two list will be used as the criterion for the following scoring-based frame selection.

% \input{fig/query}

\subsection{Key Frame Selection from Cues}
\label{sec:key frame selection}
Given the top-$k$ temporal segments predicted by the LLM, we select one representative key frame from the video frames associated with each location, resulting in $k$ query-relevant key frames. The goal is to identify frames that are maximally informative for answering the user question while remaining robust to imperfect detections. To this end, we adopt a scoring-based frame selection strategy that leverages the LLM-predicted spatial cues, namely the key object set $O_{\text{key}}$ and the cue object set $O_{\text{cue}}$. For each frame, we aggregate detection nodes' confidence scores of objects from both sets using distinct weighting factors, reflecting their different semantic roles: key objects provide direct evidence of answer relevance, while cue objects contribute complementary contextual information. This weighted aggregation approximates query-conditioned semantic alignment under partial observability, enabling robust frame selection even when key objects are weakly detected or partially missing, and avoids the brittleness of single-object or maximum-confidence selection strategies. Finally, the user question and the \(k\) selected key frames across different locations and viewpoints are passed to the VLM for question answering.

% Formally, we compute a score for each frame \(f\) using the sets of \textbf{key objects} \(O_{\text{key}}\) and \textbf{cue objects} \(O_{\text{cue}}\):  

% \[
% \text{Score}(f) \;=\; 
% \sum_{d \in D_{f}} 
% \begin{cases}
% s(d), & \text{if } d \in O_{\text{key}}, \\[6pt]
% \lambda \cdot s(d), & \text{if } d \in O_{\text{cue}}, \\[6pt]
% 0, & \text{otherwise}.
% \end{cases}
% \]

% where \(D_{f}\) is the set of all detections in frame \(f\), \(s(d)\) is the confidence score of detection \(d\), and \(\lambda\) is a weighting factor that down-weights cue objects relative to key objects which we we set to 0.1.

%% file: fig/query.tex
\begin{figure*}[t]
    \centering
    \includegraphics[width=\linewidth]{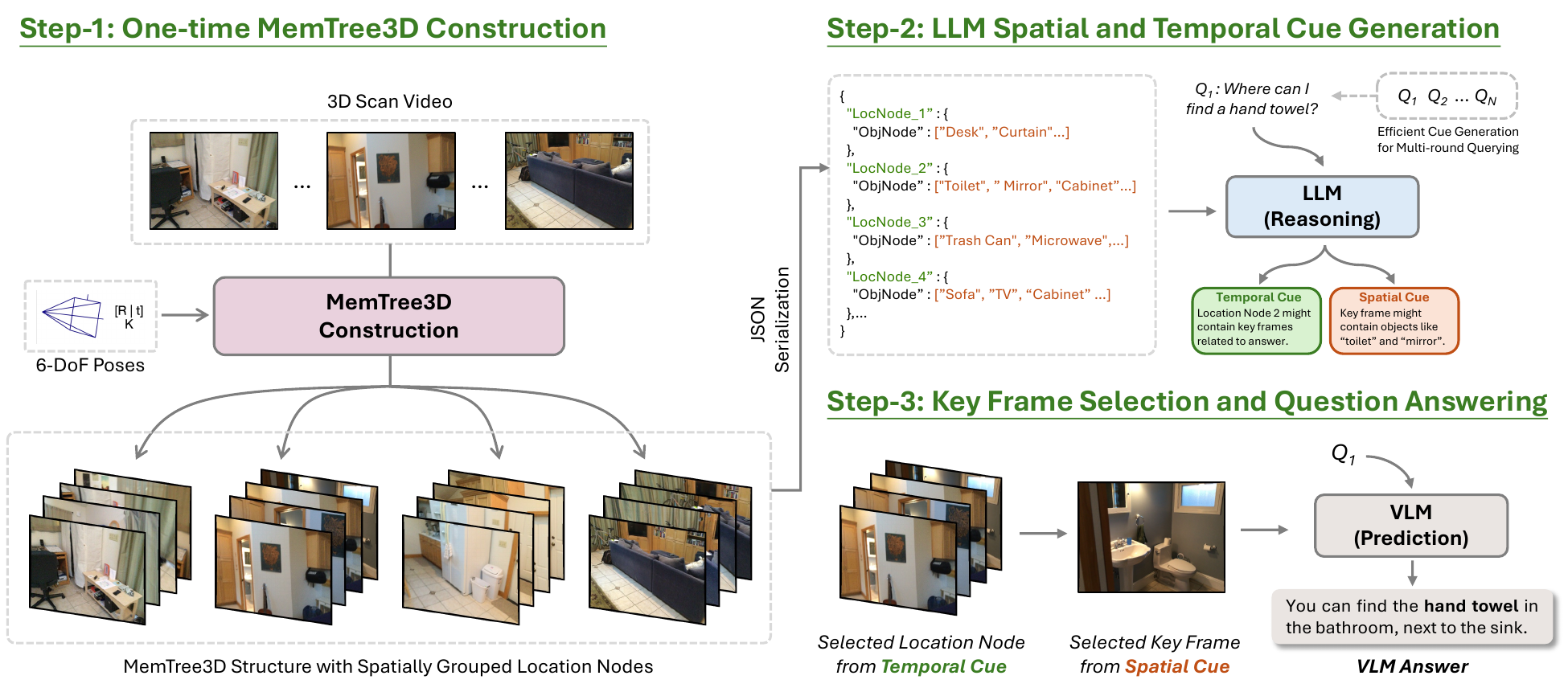}
    \caption{\textbf{Method Overview.} Step-1: A compact MemTree3D representation is constructed once from the 3D scan video using 6-DoF camera poses to spatially group frames into location nodes. Step 2: The serialized MemTree3D structure is provided to an LLM, which performs reasoning to generate temporal and spatial cues for locating relevant scene regions. Step 3: Guided by these cues, informative key frames are retrieved and passed to a VLM to produce the final answer.}
    \label{fig:query}
\end{figure*}

%% file: fig/method.tex
\begin{figure*}[t]
    \centering
    \includegraphics[width=\linewidth]{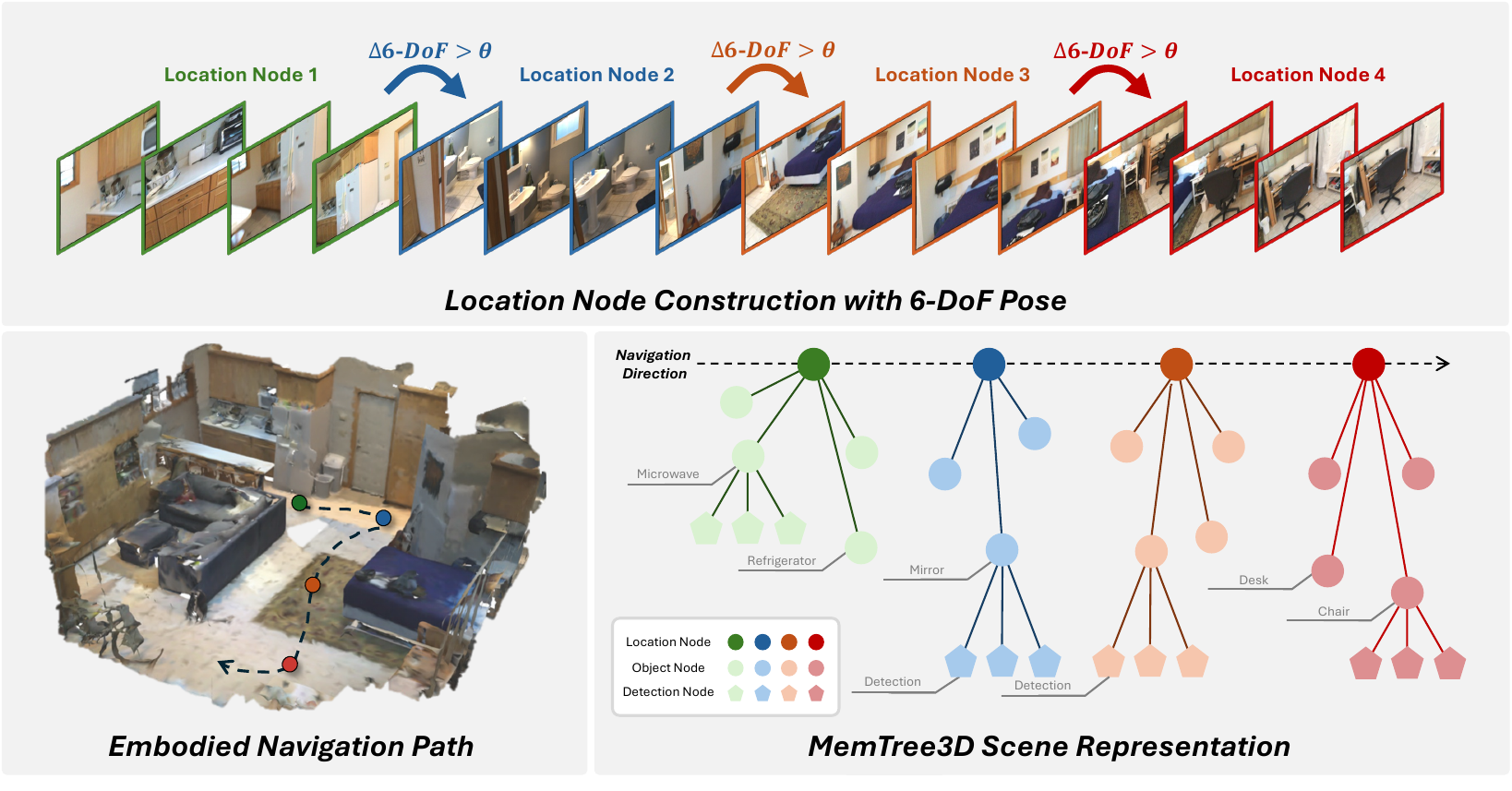}
    \caption{Our proposed \model~construction process. During the embodied navigation, the camera 6-DoF poses are used to construct the location node, each location node contains multi-level 3D scene information for later LLM key frame selection.}
    \label{fig:method}
\end{figure*}

%% file: alg/location.tex
\begin{algorithm}[t]
\caption{LocNode Construction}
\label{alg:location}
\SetAlgoNlRelativeSize{0}
\SetKw{KwParam}{\textbf{Parameters:}}
\KwIn{Poses $P[0{:}N{-}1]$, Detections $D[0{:}N{-}1]$}
\KwOut{Location nodes $\mathcal{L}$}

\setcounter{AlgoLine}{0}
$\mathcal{L} \gets \emptyset$, \quad $loc\_id \gets 0$\;
$P_{\text{prev}} \gets P[0]$\;
$d \gets \emptyset$ \tcp*[f]{Detection buffer}

\For{$t \gets 0$ \KwTo $N{-}1$}{
  $P_t \gets P[t]$\;
  $d \gets d \cup \{D[t]\}$\;
  $t_{\Delta} \gets \Delta\text{Translation}(P_{\text{prev}}, P_t)$\;
  $r_{\Delta} \gets \Delta\text{Rotation}(P_{\text{prev}}, P_t)$\;
  \If{$t_{\Delta} > T_{\text{thres}}$ \textbf{or} $r_{\Delta} > R_{\text{thres}}$}{
    \tcp{Create new location node from buffered detections}
    $\mathcal{L} \gets \mathcal{L} \cup \textbf{LocNode}(loc\_id, d)$\;
    $loc\_id \gets loc\_id + 1$\;
    $d \gets \emptyset$, \quad $P_{\text{prev}} \gets P_t$\;
  }
}

% \tcp{Append the final buffered segment, if any}
% \If{$d \neq \emptyset$}{
%   $\mathcal{L} \gets \mathcal{L} \cup \textbf{LocNode}(loc\_id, d)$\;
% }

\Return $\mathcal{L}$\;
\end{algorithm}

%% file: sec/4_experiments.tex
\section{Experiments}
\subsection{Implementation Details}
We use YOLO-World~\cite{yoloworld} and BoT-SORT~\cite{aharon2022bot} for \model~construction, following prior work~\cite{3dmem} by adopting the ScanNet-200~\cite{scannet} common indoor object categories. Additional implementation details are provided in the appendix. The tree construction runs online and in real-time (\textbf{25+ FPS}) on a single GPU, and can be easily integrated into embodied agents during the observation collection stage, in contrast to existing methods that rely on multiple vision foundation models~\cite{gu2024conceptgraphs,conceptfusion}. We set $T_{\text{thres}}$ to $\texttt{1.5 m}$ and $R_{\text{thres}}$ to $\texttt{45\textdegree}$ for all experiments, and fix the key-to-cue object weighting ratio to $10{:}1$, without careful fine-tuning. While per-scene parameter tuning could further improve temporal segmentation and spatial retrieval, we adopt a unified setting across all scenes to better reflect real-world deployment, where the number of \texttt{LocNode} adapts automatically to scene scale. The LLM selects top $k$ locations from \model~after reasoning, which we default $k$ to 3, with ablation study on different value of $k$ in Fig~\ref{fig:top_k}.

% Also note that we keep our MemTree3D design from merging the \texttt{LocNode} even when the observation revisits the same location. This preserves the temporal dimension of the memory that more closely reflects real-world settings. Furthermore, the resulting growth in JSON size remains negligible in practice compared to the computational savings gained from our approach. We conducted all the experiments on a V100 GPU.

\subsection{Benchmark}

We evaluate on OpenEQA~\cite{openeqa}, ScanQA~\cite{scanqa} and SQA3D~\cite{sqa3d}. OpenEQA is a recent Embodied Question Answering~(EQA) benchmark focusing on spatial understanding and embodied reasoning. It contains 187 episode histories collected from ScanNet~\cite{scannet} and HM3D~\cite{hm3d}, with over 1,600 human-generated questions. Furthermore, OpenEQA adopts the automatic LLM evaluation protocol to evaluate the performance of the method. We follow the official setting and report the GPT-4~\cite{gpt4} LLM-Match score. ScanQA~\cite{scanqa} and SQA3D~\cite{sqa3d} are another two large-scale benchmarks that focus on 3D scene spatial understanding, with ScanQA contains 4,675 and SQA3D contains 3,519 QA pairs. We follow previous works~\cite{llava3d,chatscene,dtc,tosa} and evaluate on ScanQA validation and SQA3D test set using Exact Match~(EM@1). More benchmark details and statistics can be found in our appendix.

\input{tab/openeqa}

\input{tab/category}
\input{tab/scanqa_sqa3d}

\subsection{Analysis on OpenEQA}

In Table~\ref{tab:openeqa}, we compared with multiple SOTA VLMs, captioning-based Socratic method~\cite{llava}, ConceptGraphs and Sparse Voxel Map~\cite{gu2024conceptgraphs} LLM methods on the OpenEQA benchmark. We also compared with three advanced visual search and 3D memory methods using GPT-4o as VLM, including detector-based VLM Grounder~\cite{xu2024vlmgrounder}, VLM-based frame selection 3D-Mem~\cite{3dmem} and an image-text retrieval baseline with CLIP~\cite{clip} that retrieve key frames by selecting the frame with highest image-query similarity.

In Table~\ref{tab:openeqa}, we show our performance implemented with open-source VLM LLaVA-OneVision-7B and proprietary model GPT-4o on the OpenEQA benchmark. Our methods largely boost the performance of existing VLMs, with \textbf{5.8\%} accuracy gain for LLaVA-One-Vision, and \textbf{17.4\%} over GPT-4o under the same number of input frame. Moreover, we achieve stronger performance compared with existing visual search methods using the same VLM GPT-4o. 

In Table~\ref{tab:category}, we report our category-level performance on OpenEQA. Integrating \model~consistently improves uniform sampling GPT-4o by more than 10\% across all question categories. Although MemTree3D adopts an object-centric representation, its role is to retrieve informative key frames rather than perform fine-grained recognition. The final disambiguation of object attributes, states, and subtle visual details is handled by the VLM using raw visual evidence from the selected frames.

\vspace{-10pt}

\subsection{Analysis on ScanQA and SQA3D}
On ScanQA and SQA3D~(Table~\ref{tab:scannet}), we compared with 3D fine-tuned models, including task-specific models~\cite{scan2cap,scanrefer,clipbert,scanqa} and 3D understanding LLMs~\cite{3d-vista,3dllm,3dvlp-jin,leo,chatscene}. In Table~\ref{tab:scannet}, our method brings accuracy improvements and achieve strong zero-shot performance compared with existing 3D fine-tuned models and 2D VLM. However, we notice the performance gain on ScanQA and SQA3D is rather moderate compared with the large performance improvements on OpenEQA, which is likely caused by the smaller scene size in ScanNet, where the uniform sampling can already achieve decent visual coverage. This observation can also be justified by the smaller improvement on the ScanNet subset~(avg. size $82.6~m^3$) compared to the HM3D subset~(avg. size $556.0~m^3$) in OpenEQA, which suggests our method is more effective in a more challenging and larger 3D scene setting.

\subsection{Ablation Studies}
\noindent{\textbf{Does the number of frames affect performance?}}
We evaluate the impact of using a larger $k$ value in Fig~\ref{fig:top_k}. Our \model~outperforms uniform sampling across different frame usage, highlighting the effectiveness of our method. Moreover, the performance of GPT-4o when using frames selected by \model~consistently improves as more frames are included, due to access to richer visual information from diverse spatial locations in the scene.

\input{fig/ablation}

\input{tab/ablation_qwen2}
\input{tab/ablation_det_fail_module}

\noindent{\textbf{How efficient is \model?}}
To demonstrate that our proposed paradigm is more efficient than visual search key frame selection, we compare our proposed \model~Frame Selection~(\model~FS) with the Detector-based Frame Selection~(Detector-based FS) method~\cite{xu2024vlmgrounder} in Fig~\ref{fig:runtime}. We use the open source LLM Qwen3~\cite{yang2025qwen3} for both methods and run them on the same V100 GPU for fair comparison and reproducibility. The runtime of Detector-based FS increases with the number of video frames, while the runtime of \model~FS remains relatively stable, with at least 69.2\% runtime speed up compared to detector-based FS, highlighting the efficiency of our LLM key frame querying paradigm. The reported runtime is measured on OpenEQA, starting from the moment the system receives the user query. Furthermore, we compare the latency of \model~and Detector-based FS in multi-round user query scenario in Fig~\ref{fig:multi-round}, the latency gap highlights the better efficiency of our \model~key frame selection paradigm compared with existing visual search approach. This efficiency stems directly from our design, which bypasses extensive per-query visual search and thereby minimizes computation.

\noindent{\textbf{Can \model~work well with open source models?}}
In Table~\ref{tab:ablation_llm}, we conduct experiments and investigate the performance of various open source and proprietary LLM and VLM combinations with our \model. We found that switching proprietary LLM GPT-4o to open source smaller scale LLM Qwen3-4B and Qwen3-8B~\cite{yang2025qwen3} and VLM LLaVA-OneVision-7B~\cite{llava-onevision} does not lead to notable performance degradation, with LLM-Match score on OpenEQA outperforms uniform sampling proprietary VLM GPT-4o.

\noindent{\textbf{Ablation on Location Node construction strategy.}} We present an ablation study of different \texttt{LocNode} construction strategies in Table~\ref{tab:ablation_module}. The uniform construction strategy partitions Location Nodes using a fixed temporal window without considering camera motion, often grouping frames from distinct viewpoints into the same node. Increasing the node construction frequency by reducing the temporal window (from 90 to 30 frames) leads to improved performance, but still underperforms the 6-DoF–based construction, which explicitly leverages camera motion to segment the 3D scene into semantically coherent nodes, yielding a significant improvement of 8.7\% over uniform sampling.

\noindent{\textbf{Is \model~robust to perception failure?}}
To evaluate the robustness of our framework under perception failures during \model~construction, we analyze the performance gap with respect to \textbf{query object presence}, as reported in Table~\ref{tab:ablation_det}. We apply a heuristic string-matching strategy to determine whether object class names present in the constructed scene representation appear in the user question. On OpenEQA, our method achieves an LLM-Match score of 65.9 in the absence of query object presence, representing only a 3.1-point drop compared to cases where query objects are present. 

% Manual verification further shows that the ratio of full score answers without correct key-frame retrieval is only \textbf{0.25\%}, underscoring the critical role of effective key-frame selection. Together, these results highlight our LLM-guided key-frame selection does not over-rely on visual perception results and further distinguishes our approach from existing visual search and scene graph–based methods.

% \input{tab/ablation_det_fail}
% \input{fig/ablation_clutter}
\input{fig/qual}

% \paragraph{Does 3D scene complexity affect performance?}
% A critical evaluation for our framework is its robustness in complex 3D scenes with significant clutter. To quantify scene complexity, we introduce a clutter level, defined as the total number of unique objects divided by the number of \texttt{LocNode} in a 3D scene. A higher clutter level indicates more objects per location, thereby increasing the difficulty of selecting the correct location and frame. As shown in Fig~\ref{fig:clutter}, our method's LLM-Match score on OpenEQA benchmark does not degrade significantly as the clutter level increases. This result demonstrates that our framework's LLM-based selection mechanism is not confounded by distractor objects, enabling it to intelligently identify the most salient locations and keyframes for efficient and accurate question answering.

% \noindent{\textbf{Frame Storage Feasibility and Efficiency.}}
% One potential concern can be raised is the memory requirement introduced by storing the RGB frames for retrieval and VLM inference. However, we emphasize that the benefits of our method substantially outweigh this cost. First, the VRAM requirement of running the VLM on a large number of frames imposes far greater cost than storing frames on conventional storage such as SSD/HDD. Second, modern embodied devices can easily accommodate extended video data such as several days of high-resolution footage. Therefore, the challenge lies in the efficient retrieval of relevant frames, which is precisely addressed by our memory-efficient mechanism for 3D question answering task.

\subsection{Qualitative Results} We present qualitative results in Fig~\ref{fig:qual}, which covers multiple question types such as location reasoning, multi-view selection, and spatial understanding.

\noindent{\textbf{Location Reasoning.}} When query-relevant key objects are missing from the \model~representation due to perception failures, our pipeline uses the LLM to infer likely locations from the objects observed at each location.

\noindent{\textbf{Multi-view Selection.}} For questions requiring multiple viewpoints, our method retrieves frames from different location to provide comprehensive visual context.

\noindent{\textbf{Spatial Understanding and Visual Recognition.}} By selecting visually informative frames, our key-frame selection improves spatial reasoning and fine-grained visual recognition, resulting in more accurate answers.

%% file: tab/openeqa.tex
\begin{table*}[t]
\centering
\caption{Performance comparison of our \model~with other SoTA methods on OpenEQA, including the performance on ScanNet and HM3D subset. Performance of \model~with different number of used frame can be found in ablation study.}
\resizebox{0.98\linewidth}{!}{
\begin{tabular}{l c c c c}
\toprule
\multicolumn{2}{l}{} & \multicolumn{3}{c}{LLM-Match} \\
\cmidrule{3-5}
\textbf{Method} & \textbf{Avg. Frame} & \makebox[5em][c]{\textbf{ScanNet}} & \makebox[5em][c]{\textbf{HM3D}} & \makebox[5em][c]{\textbf{ALL}} \\
\midrule
\textbf{\textit{Open-source VLM}} & \\
\rowcolor{gray!15} Video-LLaMA~\cite{videollama} & 8 & 20.1 & 19.8 & 20.0 \\
Video-ChatGPT~\cite{videochatgpt} & 100 & 32.9 & 30.4 & 32.1 \\
\rowcolor{gray!15} LLaMA-2 w/ Concept Graph~\cite{openeqa} & 50 & 31.0 & 24.2 & 28.7 \\
LLaMA-2 w/ Sparse Voxel Map~\cite{openeqa} & 50 & 36.0 & 30.9 & 34.3 \\
\rowcolor{gray!15} LLaMA-2 w/ LLaVA-1.5 caption~\cite{openeqa} & 50 & 39.6 & 31.1 & 36.8 \\
Qwen-2.5-VL-7B~\cite{bai2025qwen25vl} & 12 & 49.4 & 40.2 & 46.2 \\
\rowcolor{gray!15} Video-LLaMA2~\cite{videollama2} & 16 & 50.1 & 47.5 & 49.2 \\
LLaVA-3D~\cite{llava3d} & 32 & -- & -- & 53.2 \\
% \rowcolor{gray!15} LLaVA-OneVision-7B~\cite{llava-onevision} & 12 & 58.5 & 51.6 & 56.2\\
\midrule
\textbf{\textit{Closed-source VLM}} & \\
% LLaMA-2 70B$^\dagger$ & -- & 28.3 \\
% GPT4$^\dagger$ & -- & 33.5 \\
\rowcolor{gray!15} Claude-3 Opus~\cite{anthropic2024claude3} & 20 & -- & -- & 36.3 \\
Gemini 1.0 Pro Vision~\cite{gemini} & 15 & -- & -- & 44.9 \\
\rowcolor{gray!15} Claude-3.5 Sonnet~\cite{anthropic2024claude3} & 20 & -- & -- & 48.7 \\
GPT-4V w/ Concept Graph~\cite{openeqa} & 50 & 37.8 & 34.0 & 36.5 \\
\rowcolor{gray!15} GPT-4V w/ Sparse Voxel Map~\cite{openeqa} & 50 & 40.9 & 35.0 & 38.9 \\
GPT-4V w/ LLaVA-1.5 caption~\cite{openeqa} & 50 & 45.4 & 40.0 & 43.6\\
% GPT-4V & 15 & & & 54.6 \\
\rowcolor{gray!15} GPT-4V~\cite{gpt4} & 50 & 57.4 & 51.3 & 55.3 \\
% GPT-4o & 3 & -- & -- & 48.1\\
% GPT-4o~\cite{4o} & 6 & 58.2 & 49.6 & 55.3\\
GPT-4o~\cite{4o} & 12 & 63.9 & 58.4 & 62.0\\
% GPT-4V & 15 / 50 & 54.6 / 55.3 \\
% GPT-4o & 3 / 6 / 12 & 48.1 / 55.3 / 62.0\\
% \midrule
% \textit{Token Merging / Pruning Method}\\
% LLaVA-OneVision-7B w/ ToMe~\cite{tome} & - & - & -- & 48.3\\
% LLaVA-OneVision-7B w/ ToSA~\cite{tosa} & - & - & -- & 49.5\\
% LLaVA-OneVision-7B w/ DTC~\cite{dtc} & - & - & -- & 52.5\\
\midrule
\textbf{\textit{Visual Search / 3D Memory Method}}\\ 
\rowcolor{gray!15} CLIP Retrieval~\cite{clip} & 3 & 55.1 & 41.9 & 50.6\\
3D-Mem~\cite{3dmem} & 3.1 & -- & -- & 57.2\\
\rowcolor{gray!15} VLM-Grounder~\cite{xu2024vlmgrounder} & 6 & 65.1 & 58.8 & 63.0\\
% \rowcolor{gray!15} AKS~\cite{adaptive} & 3 & -- & -- & --\\
% T*~\cite{tstar} & 3 & -- & -- & --\\
\midrule
\textbf{\textit{Ours~(with zero-shot 2D VLM)}} & \\
\rowcolor{gray!15} LLaVA-OneVision-7B~\cite{llava-onevision} & 3 & 52.5 & 42.9 & 49.2\\
\textbf{\model}~(w/ LLaVA-OneVision-7B) & 3 & 57.9~(+5.4) & 49.3~(+6.4) & 55.0~(+5.8)\\
\rowcolor{gray!15} GPT-4o~\cite{4o} & 3 & 55.8 & 37.1 & 49.4 \\
\textbf{\model}~(w/ GPT-4o) & 3 & \textbf{69.4~(+13.6)} & \textbf{61.7~(+24.6)} & \textbf{66.8~(+17.4)} \\ % use 6-DOF pose
\bottomrule
\end{tabular}
\label{tab:openeqa}
}
\end{table*}

%% file: tab/category.tex
\begin{table*}[t]
\centering
\caption{Category-level performance on OpenEQA.}
\resizebox{0.98\linewidth}{!}{
\begin{tabular}{lccccccc}
\hline
\multicolumn{8}{c}{\hspace{9em}\textbf{EQA Category}} \\
\cline{2-8}
\multirow{2}{*}{Method} & object & object & attribute & spatial & object state & functional & world\\
 & recognition & localization & recognition & understanding & recognition & reasoning & knowledge\\
\hline
GPT-4o & 54.0 & 36.9 & 58.3 & 36.8 & 54.6 & 53.4 & 50.2 \\
w/ MemTree3D & 64.6 & 58.1 & 68.9 & 52.5 & 75.4 & 73.6 & 72.3\\
\hline
$\mathrm{\Delta}$
& +10.6 
& +21.2 
& +10.6 
& +15.7 
& +20.8 
& +20.2 
& +22.1 \\

\hline

\end{tabular}
\label{tab:category}
}
\end{table*}

%% file: tab/scanqa_sqa3d.tex
\begin{table}[h]
\centering
% \small
\caption{EM@1 comparison on ScanQA and SQA3D.}
% \caption{Exact Match~(EM@1) and Refined Exact Match~(R-EM@1) performance on ScanQA and SQA3D.}
\resizebox{0.8\linewidth}{!}{
\begin{tabular}{l c c c}
\toprule
\multicolumn{2}{l}{} & \multicolumn{2}{c}{EM@1} \\
\cmidrule{3-4}
\textbf{Method} & \textbf{Avg. Frame} & \makebox[2.5em][c]{\textbf{ScanQA}} & \makebox[2.5em][c]{\textbf{SQA3D}} \\
% Method & Avg. Frame & ScanQA & SQA3D\\
\midrule
\textbf{\textit{3D Fine-tuned Model}} \\
\rowcolor{gray!15} Scan2Cap~\cite{scan2cap} & -- & -- & 41.0 \\
ScanRefer+MCAN~\cite{scanrefer} & -- & 18.6 & -- \\
\rowcolor{gray!15} ClipBERT~\cite{clipbert} & -- & -- & 43.3 \\
ScanQA~\cite{scanqa} & -- & 21.1 & 47.2 \\
\rowcolor{gray!15} 3D-VisTA~\cite{3d-vista} & -- & 22.4 & 48.5 \\
3D-LLM~\cite{3dllm} & -- & 20.5 & 48.1 \\
\rowcolor{gray!15} 3D-VLP~\cite{3dvlp-jin} & -- & 21.6 & -- \\
LEO~\cite{leo} & -- & 24.5 & 50.0 \\
% \rowcolor{gray!15} Scene-LLM~\cite{scenellm} & -- & 27.2 & 54.2 \\
\rowcolor{gray!15} ChatScene~\cite{chatscene} & -- & 21.6 & 54.6 \\
% \rowcolor{gray!15} LLaVA-3D~\cite{llava3d} & -- & 27.0 & 55.6 \\
% \textit{2D Models} \\
% LLaVA-NeXT-Video~\cite{llavanext} & -- & 18.7 & 34.2 \\
% LLaVA-Video~\cite{llava-video} & -- & -- & 48.5 \\
% LLaVA-OneVision-7B~\cite{llava-onevision} & -- & 27.2 & 51.4 \\
% GPT-4V~\cite{gpt4} & -- & -- & -- \\
% Gemini~\cite{gemini} & -- & -- & -- \\
% Claude~\cite{anthropic2024claude3} & -- & -- & -- \\
\midrule
\textbf{\textit{VLM Method}} \\
\rowcolor{gray!15} Agent3D-zero~\cite{zhang2024agent3d} & 24 & 17.5 & -- \\
LLaVA-Next-Video~\cite{llavanext} & 32 & 18.7 & 34.2 \\
\rowcolor{gray!15} VideoChat2~\cite{videochat2} & 16 & 19.2 & 37.3 \\
% \midrule
% \textbf{\textit{Ours~(with zero-shot 2D VLM)}} & \\
LLaVA-OneVision-7B~\cite{llava-onevision} & 3 & 25.1 & 46.2 \\
% \rowcolor{gray!15} LLaVA-OneVision-7B~\cite{llava-onevision} & 3 & 25.1~(42.6) & 46.2~(50.5) \\
% \wboldmodel & 3 & \textbf{27.1~(+2.0)} & \textbf{49.4~(+3.2)} \\
% \wboldmodel & 3 & \textbf{28.0~(46.8)} & \textbf{49.6~(52.9)} \\
% \wboldmodel & 3 & \textbf{28.0} & \textbf{49.6} \\
\rowcolor{gray!15} \boldmodel~(w/ LLaVA-OneVision-7B) & 3 & \textbf{28.0~(+2.9)} & \textbf{49.6~(+3.4)} \\
\bottomrule
\end{tabular}
}
\label{tab:scannet}
\end{table}

%% file: fig/ablation.tex
% \begin{figure*}[!t]
%     \centering
%     \begin{minipage}{0.32\linewidth}
%         \centering
%         \includegraphics[width=\linewidth]{fig/top_k_5.pdf}
%         \caption{OpenEQA performance comparison with uniform sampling using different number of frames.}
%         \label{fig:top_k}
%     \end{minipage}
%     \hfill
%     \begin{minipage}{0.32\linewidth}
%         \centering
%         \includegraphics[width=\linewidth]{fig/time_4.pdf}
%         \caption{Runtime comparison with Detector-based FS method. Runtime is measured after receiving question.}
%         \label{fig:runtime}
%     \end{minipage}
%     \hfill
%     \begin{minipage}{0.32\linewidth}
%         \centering
%         \includegraphics[width=\linewidth]{fig/multi-round-figure.pdf}
%         \caption{Runtime comparison with Detector-based frame sampling method when processing multi-round query.}
%         \label{fig:mr-query}
%     \end{minipage}
% \end{figure*}

% \begin{figure}[!t]
%     \centering
    
%     \begin{minipage}{0.6\linewidth}
%     \begin{minipage}{0.48\linewidth}
%         \centering
%         \includegraphics[width=\linewidth]{fig/top_k_4.pdf}
%         \caption{OpenEQA performance comparison with uniform sampling using different number of frames.}
%         \label{fig:top_k}
%     \end{minipage}
%     \hfill
%     \begin{minipage}{0.48\linewidth}
%         \centering
%         \includegraphics[width=\linewidth]{fig/time_4.pdf}
%         \caption{Runtime comparison with Detector-based FS method. Runtime is measured after receiving question.}
%         \label{fig:runtime}
%     \end{minipage}
%     \end{minipage}
% \end{figure}

\begin{figure}[!t]
    \centering
    \begin{minipage}{0.98\linewidth}
        \centering
        \begin{minipage}{0.32\linewidth}
            \centering
            \includegraphics[width=\linewidth]{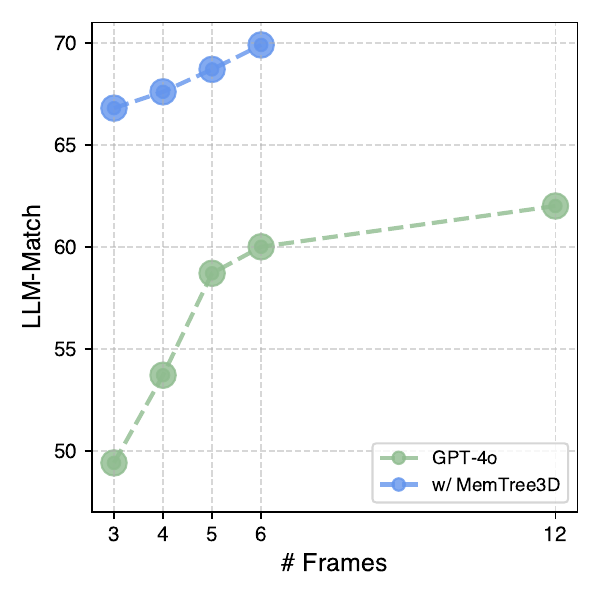}
            \caption{OpenEQA results comparison with uniform sampling using different number of frames.}
            \label{fig:top_k}
        \end{minipage}
        \hfill
        \begin{minipage}{0.32\linewidth}
            \centering
            \includegraphics[width=\linewidth]{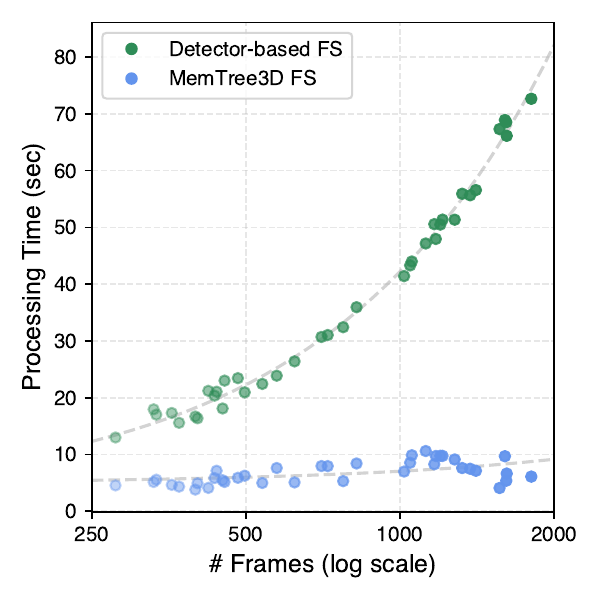}
            \caption{Runtime comparison with Detector-based FS method. Measured after received question.}
            \label{fig:runtime}
        \end{minipage}
        \hfill
        \begin{minipage}{0.32\linewidth}
            \centering
            \includegraphics[width=1.01\linewidth]{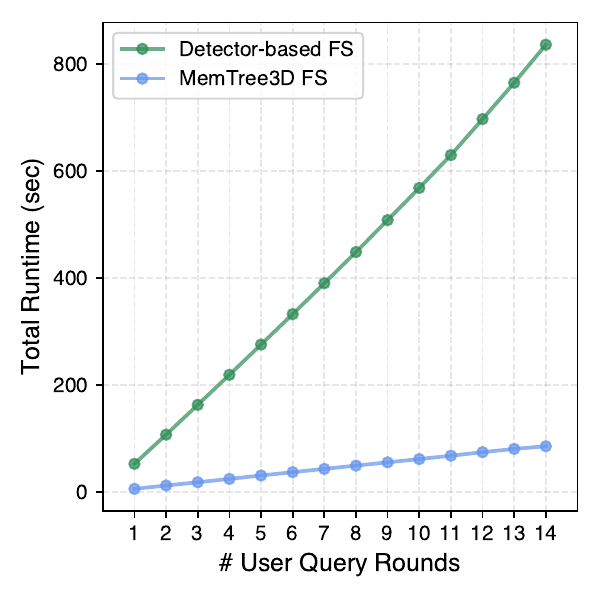}
            \caption{Runtime comparison with Detector-based FS method when processing multi-round query.}
            \label{fig:multi-round}
        \end{minipage}
    \end{minipage}
\end{figure}

%% file: tab/ablation_qwen2.tex
\begin{table}[!t]
\centering
\caption{Ablation on performance using different LLM/VLM combinations.}
\resizebox{0.8\linewidth}{!}{
\footnotesize
\begin{tabular}{ccccc}
\toprule
% \multicolumn{2}{l}{} & \multicolumn{3}{c}{LLM-Match} \\
% \cmidrule{3-5}
LLM & VLM & \makebox[4em][c]{ScanNet} & \makebox[4em][c]{HM3D} & \makebox[4em][c]{All} \\
\midrule
% \textit{Uniform Sampling}\\
\textcolor{gray}{--} & \textcolor{gray}{GPT-4o~(Uniform Sampling)} & \textcolor{gray}{55.8} & \textcolor{gray}{37.1} & \textcolor{gray}{49.4} \\
% \midrule
% \textit{Ours}\\
Qwen3-4B & LLaVA-OneVision-7B & 54.9 & 43.8 & 51.1\\
Qwen3-8B & LLaVA-OneVision-7B & 56.8 & 46.0 & 53.1\\
GPT-4o & LLaVA-OneVision-7B & 57.9 & 49.3 & 55.0\\
GPT-4o & GPT-4o   & 69.4 & 61.7 & 66.8\\
\bottomrule
\end{tabular}
}
\label{tab:ablation_llm}
\end{table}

%% file: tab/ablation_det_fail_module.tex
\begin{table}[t]
\centering
\small
\begin{minipage}{0.45\linewidth}
    \centering
    \caption{Ablation on LocNode construction strategy.}
    \resizebox{0.98\linewidth}{!}{
\begin{tabular}{l @{\hspace{1em}} c @{\hspace{1em}} c}
\toprule
Method & Avg. Frame Per Node & LLM-Match \\
\midrule
Uniform & 90 & 33.0 \\
Uniform & 30 & 40.4 \\
6-DoF & 36.3 & \textbf{49.1} \\
\bottomrule
\end{tabular}
    }
    \label{tab:ablation_module}
\end{minipage}
\hfill
\begin{minipage}{0.53\linewidth}
    \centering
    \caption{System robustness analysis under perception failure.}
    \resizebox{0.9\linewidth}{!}{
\begin{tabular}{c @{\hspace{1.2em}} c @{\hspace{1.2em}} c @{\hspace{1.2em}} c}
\toprule
Query Object Presence & ScanNet & HM3D & All \\
\midrule
\cmark & 71.4 & 64.3 & 69.0 \\
\xmark & 67.9 & 62.1 & 65.9 \\
\bottomrule
\end{tabular}
    }
    \label{tab:ablation_det}
\end{minipage}
\end{table}

%% file: fig/qual.tex
\begin{figure*}[t]
    \centering
    \includegraphics[width=\linewidth]{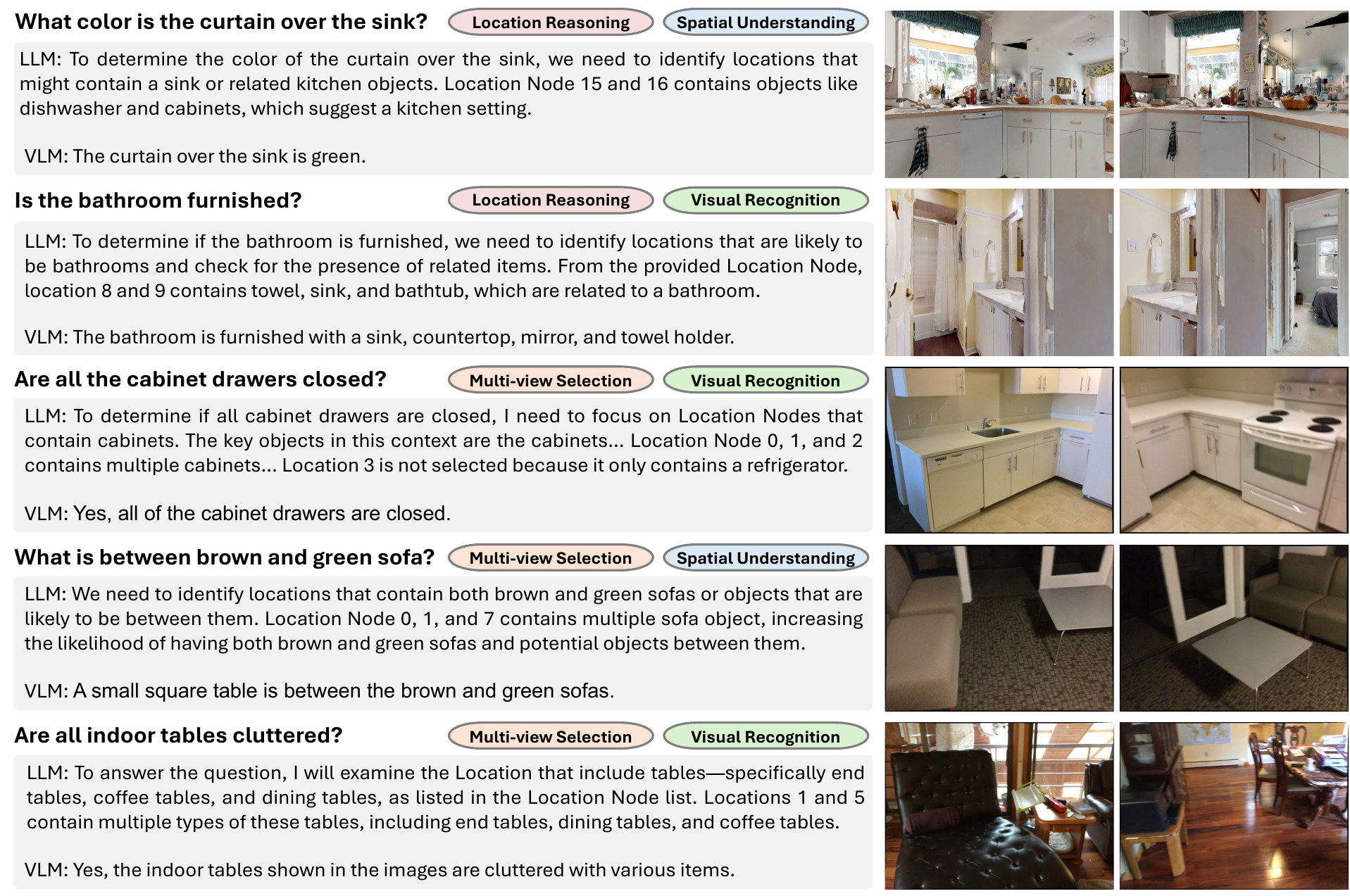}
    \caption{\textbf{Qualitative results}. We present the top two selected key frames by our method, as well as the corresponding LLM location selection and VLM output.}
    \label{fig:qual}
\end{figure*}

%% file: sec/6_conclusion.tex
\section{Conclusion}
We introduce the \model~-guided key-frame selection paradigm for efficient and accurate 3D question answering. By constructing a compact 3D scene representation, \model~enables LLM-based reasoning to select key frames without exhaustive visual search. Experiments on OpenEQA, ScanQA, and SQA3D show consistent improvements across both open-source and proprietary VLMs using fewer input frames, demonstrating the potential of our approach for real-world embodied AI applications.

%% file: sec/X_suppl.tex
\clearpage
% \setcounter{page}{1}
% \maketitlesupplementary

\section{Appendix}
\noindent We present more details and experiment results in the supplementary material structured as follows:

\begin{itemize}
\setlength{\itemsep}{2pt}
% \item Implementation details.
\item Evaluation benchmark details.
\item LLM prompt for key frame querying.
\item More results on efficiency analysis.
\item Additional qualitative results.
\item Limitations and failure cases.
\end{itemize}

% \section{Implementation Details}

% \begin{table}[h]
% \centering
% \caption{LLM \& VLM Inference configurations}
% \label{tab:4o_params}
% \begin{tabular}{|l|c|c|c|}
% \hline
% \textbf{Parameter} & \textbf{GPT-4o} & \textbf{Qwen3} & \textbf{LLaVA-OV} \\ \hline
%     temperature & 0.0 & 0.6 & 0.0\\ \hline
% \end{tabular}
% \end{table}

\section{Evaluation Benchmarks Details}

\begin{table}[ht]
    \small
    \centering
    \centering
    \caption{Scale comparison of the evaluated 3D question answering benchmarks in our work.}
    \resizebox{0.5\linewidth}{!}{
    \begin{tabular}{ccc}
    \toprule
        Benchmark &  \# of scenes & \# of questions\\
        \midrule
        OpenEQA~\cite{openeqa} & \textbf{152} & 1,636\\
        ScanQA~\cite{scanqa} & \underline{71} & \textbf{4,306}\\
        SQA3D~\cite{sqa3d} & 67 & \underline{3,519}\\
        
    \bottomrule
    \end{tabular}
    }
    \label{tab:scale}
\end{table}

\begin{table}[ht]
    \centering
    \caption{Statistics of the ScanNet and HM3D subset from the OpenEQA. The table shows the number of scenes, questions, and the average scene size.}
    \resizebox{0.8\linewidth}{!}{
    \begin{tabular}{ccccc}
    \toprule
     Subset & \# scenes & avg. size~(\text{$m^3$}) & avg. LocNode & $\Delta$~LLM-Match\\
     \midrule
     ScanNet & 89 & 82.6 & 5.9 & +13.6\\
     HM3D & 63 & 556.0 & 20.3 & +24.6\\
     \bottomrule
    \end{tabular}

    }
    \label{tab:scene_size}
\end{table}

In this work, we evaluate on 3 different benchmarks, including OpenEQA~\cite{openeqa}, ScanQA~\cite{scanqa}, and SQA3D~\cite{sqa3d}. We provide detailed statistics of these benchmarks in Table~\ref{tab:scale}. Among these benchmarks, OpenEQA has the most 3D scenes, featuring 3D scene collected from ScanNet~\cite{scannet} and HM3D~\cite{hm3d}. And ScanQA having the largest scale with 4,306 questions.

We also compare the two subsets from OpenEQA in Table~\ref{tab:scene_size}, including the number of 3D scene, average 3D scene size, average number of constructed \texttt{LocNode}, and the performance gain when using our \model. As discussed in our main paper, we notice a larger performance gain in the HM3D subset compare with the ScanNet subset, we believe this is caused by the larger 3D scene size in HM3D. In ScanNet, the 3D scene size is much smaller, and uniform sampling serve as a strong baseline strategy to cover all the visual information from the 3D scene. However, in HM3D, the 3D scene size is much larger, and uniform sampling fails to cover all the visual details. Our proposed key frame selection strategy can address this challenges by selecting the most relevant key frame to the question, with a significant 24.6\% performance gain on HM3D subset.

\input{fig/multi-round}

\section{More detailed on efficiency analysis}
We compare the runtime efficiency of our \model~key frame selection framework against existing visual search-based approach, as shown in Fig~\ref{fig:runtime}. We also compare the efficiency analysis of multi-round query in Fig~\ref{fig:multi-round2}. Specifically, we measure the latency between receiving a user question and producing a response—an essential factor in real-world embodied AI scenarios. In such settings, an agent typically navigates in a 3D scene to collect visual observations and then answers user queries based on the visual stream. The key performance metric is the response time after receiving the user question. Visual search-based methods, such as VLM-Grounder~\cite{xu2024vlmgrounder}, involve a multi-stage pipeline after receiving a question, resulting in substantial latency. The process typically includes:

\begin{enumerate}
\item Using an LLM to parse the question and extract a list of target objects relevant to the query.
\item Running open-vocabulary object detector on all video frames using the identified target objects as text prompts.
\item Sending the subset of frames containing detected objects to a VLM for question answering.
\end{enumerate}

Among these steps, step 2 leads to large latency issue due to the extensive computational cost growing proportionally with the number of video frames. In contrast, our tree based frame selection framework pre-builds the 3D representation before receiving the user's question, instead of searching the answer in the video frames, our work searches the key frame from the constructed \model, therefore bypasses this extensive visual search process, and incurs minimal latency regardless of the video length, as illustrated in Fig~\ref{fig:runtime}.

\section{Implementation Details}

\paragraph{LLM/VLM Inference Parameters.}
We set the inference temperature of \texttt{GPT-4o} and LLaVA-OneVision-7B to 0.0, and 0.6 for Qwen3 following the default setting. The rest of the inference parameters follow the default generation configuration from their corresponding huggingface repositories and official API setting.

\paragraph{LLM Location Selection Prompt.}
We present the full prompt used for location selection with \texttt{GPT-4o} in Fig~\ref{fig:prompt}. The prompt includes a one-shot example illustrating the expected JSON response format, along with custom tags such as \texttt{<think>} and \texttt{<answer>} to guide the model's reasoning and output parsing.

\paragraph{Image-text Retrieval Baseline.}
We implement the Image-text retrieval baseline using CLIP~\cite{clip} from the \texttt{openai-clip-vit-base-patch32} checkpoint. We directly use the user question as input text and compare the image-text cosine similarity to retrieve the top $k$ images for \texttt{GPT-4o} VLM input.

% \section{Ablation Study on Location Construction}

% \input{tab/ablation_module}

% We present an ablation study of different \texttt{LocNode} construction strategies in Table~\ref{tab:ablation_module}, focusing on their impact on key frame selection performance. The experiments are conducted on the challenging HM3D subset of OpenEQA, where the goal is to select the most relevant location and frame~(top-1) for answering the question.

% We compare two \texttt{LocNode} construction methods: (1) a uniform segmentation approach that partitions the video into fixed-length intervals (30 frames), and (2) our proposed 6-DoF-based method that leverages camera pose changes to segment the scene more meaningfully. In addition, we evaluate the effect of frame selection strategies by comparing random sampling against our proposed scoring-based frame selection method.

% Results show that 6-DoF \texttt{LocNode} construction outperforms the uniform baseline by a large margin, highlighting the importance of spatial awareness in location segmentation. Furthermore, our scoring-based frame selection also provides a performance improvements over random selection, but the rather moderate performance improvements suggest the 6-DoF \texttt{LocNode} construction already partition the 3D scene with good quality, and each \texttt{LocNode} covers visually similar content. This leads to a rather moderate performance gain between using random frame selection and scoring-based frame selection strategy.

\section{Additional Qualitative Results}
We present more qualitative results in Fig~\ref{fig:supp_qual}. With more frame selection results comparison with VLM image-text  retrieval baseline method~\cite{clip}. Here, we present several failure cases of the VLM image-text retrieval baseline, including 1. VLM fails to retrieve any key frames related to question (row 1), where the key frame selection required question reasoning and location understanding. 2. VLM samples key frames from the similar location and viewpoint, and failed to retrieve correct key frame for the question (row 2-4), as direct selecting top-k frames based on similarity score can lead to over sample on the similar location and viewpoint, our \model~segments the 3D scene into multiple \texttt{LocNode} and alleviates this issue.

\section{Limitations and Failure Cases}

Despite the better efficiency and significant performance gain over existing visual search key frame selection method, our \model~still has its limitation. Our tree based key frame selection paradigm can occasionally fail in some challenging object localization question where the object is not directly detected. More specifically, when the object localization question is querying an entirely novel object that is not part of \model, the LLM during key frame querying process will make its most educated guess to select several most possible locations that the target object might appear, but this best effort of reasoning does not always yield successful object localization. We illustrate several failure cases in Fig~\ref{fig:fail}, where the LLM selects several reasonable locations but fails to retrieve the target objects.

\input{fig/prompt}
\input{fig/supp_qual}
\input{fig/fail}

%% file: fig/multi-round.tex
\begin{figure}[t]
    \centering
    \includegraphics[width=0.7\linewidth]{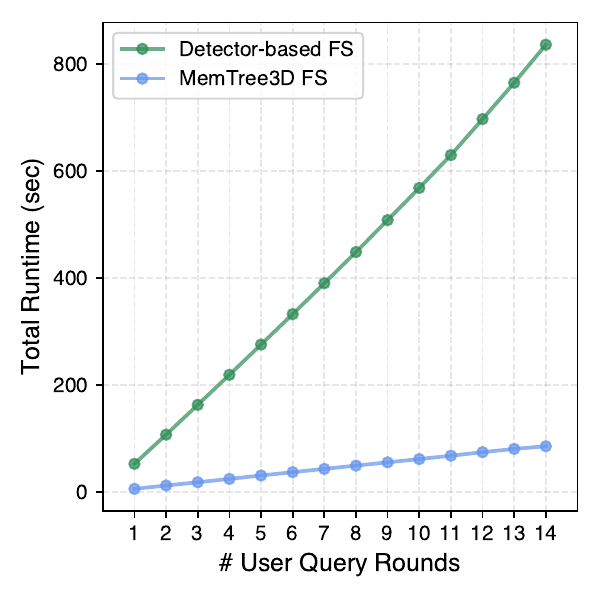}
    \caption{Runtime comparison with Detector-based frame sampling method when processing multi-round query. Our method consistently achieve efficient key frame selection compared with detector-based frame selection approach.}
    \label{fig:multi-round2}
\end{figure}

%% file: fig/prompt.tex
\begin{figure*}[!t]
    \centering
    \includegraphics[width=\linewidth]{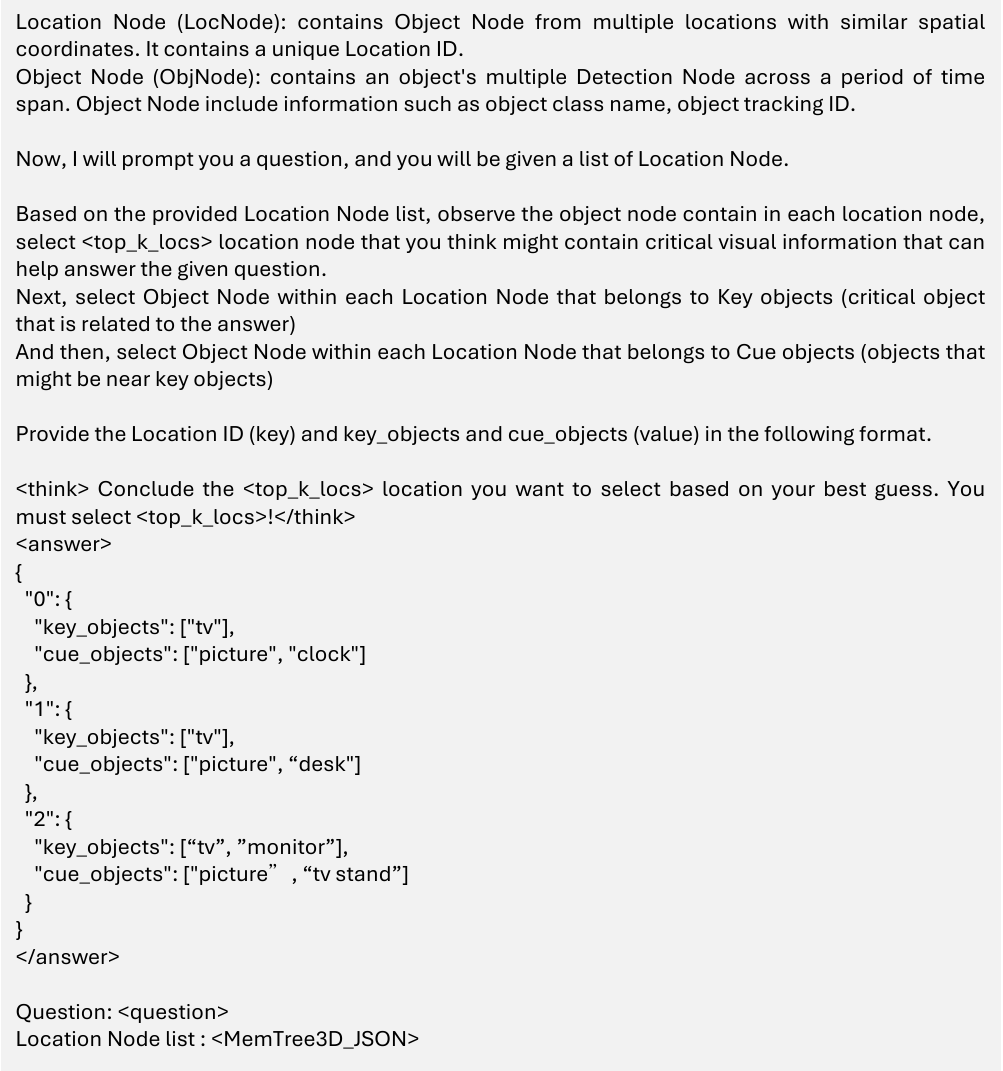}
    \caption{LLM prompt for \model~querying and location selection.}
    \label{fig:prompt}
\end{figure*}
\clearpage

%% file: fig/supp_qual.tex
\begin{figure*}[t]
    \centering
    \includegraphics[width=\linewidth]{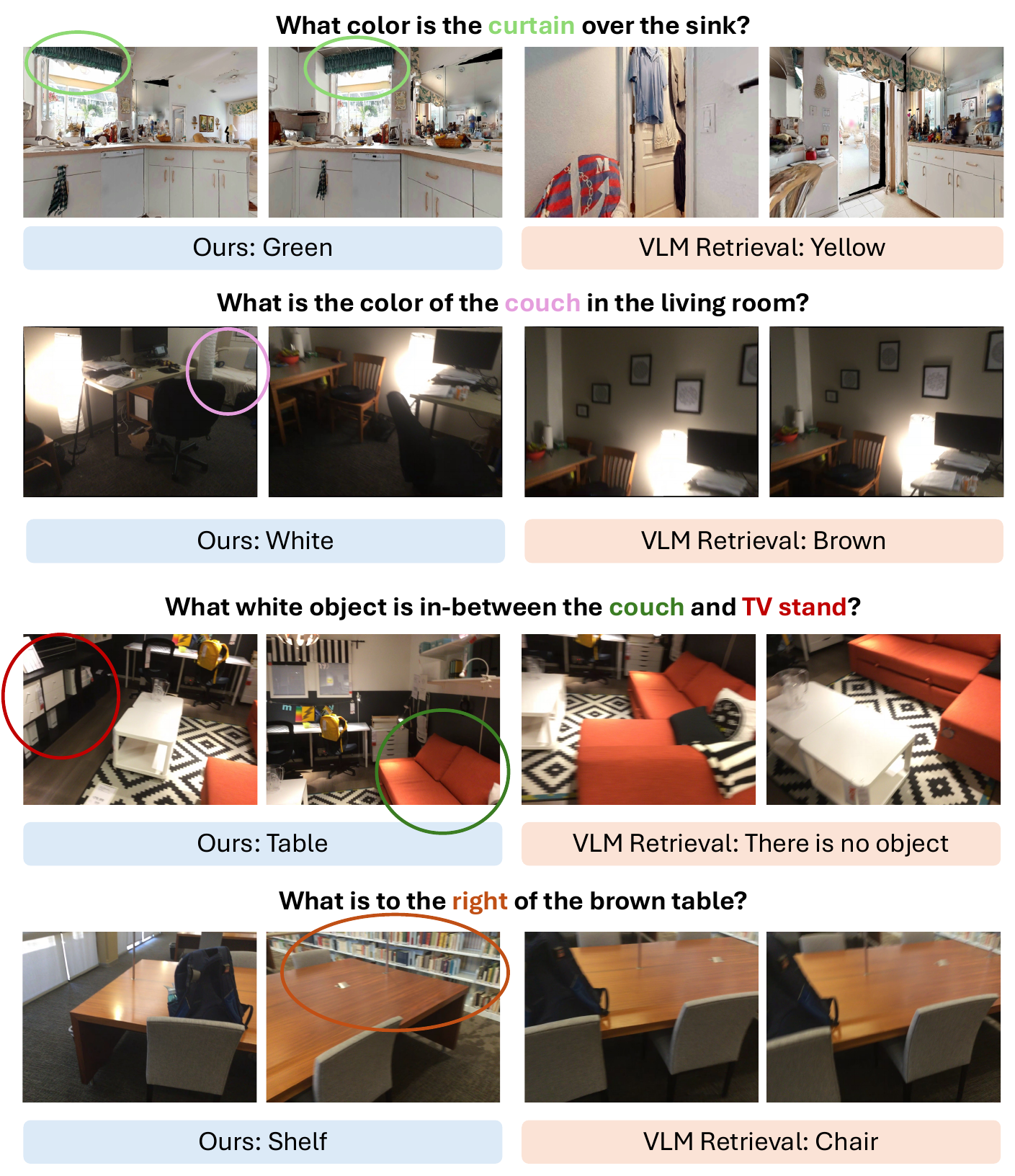}
    \caption{Additional qualitative results comparison between our \model~framework and VLM retrieval baseline. We showcase several cases, including 1. VLM fails to retrieve any key frames related to question (row 1). 2. VLM samples key frames from the similar location and viewpoint, and failed to retrieve correct key frame for the question (row 2-4).}
    \label{fig:supp_qual}
\end{figure*}

%% file: fig/fail.tex
\begin{figure*}[t]
    \centering
\includegraphics[width=\linewidth]{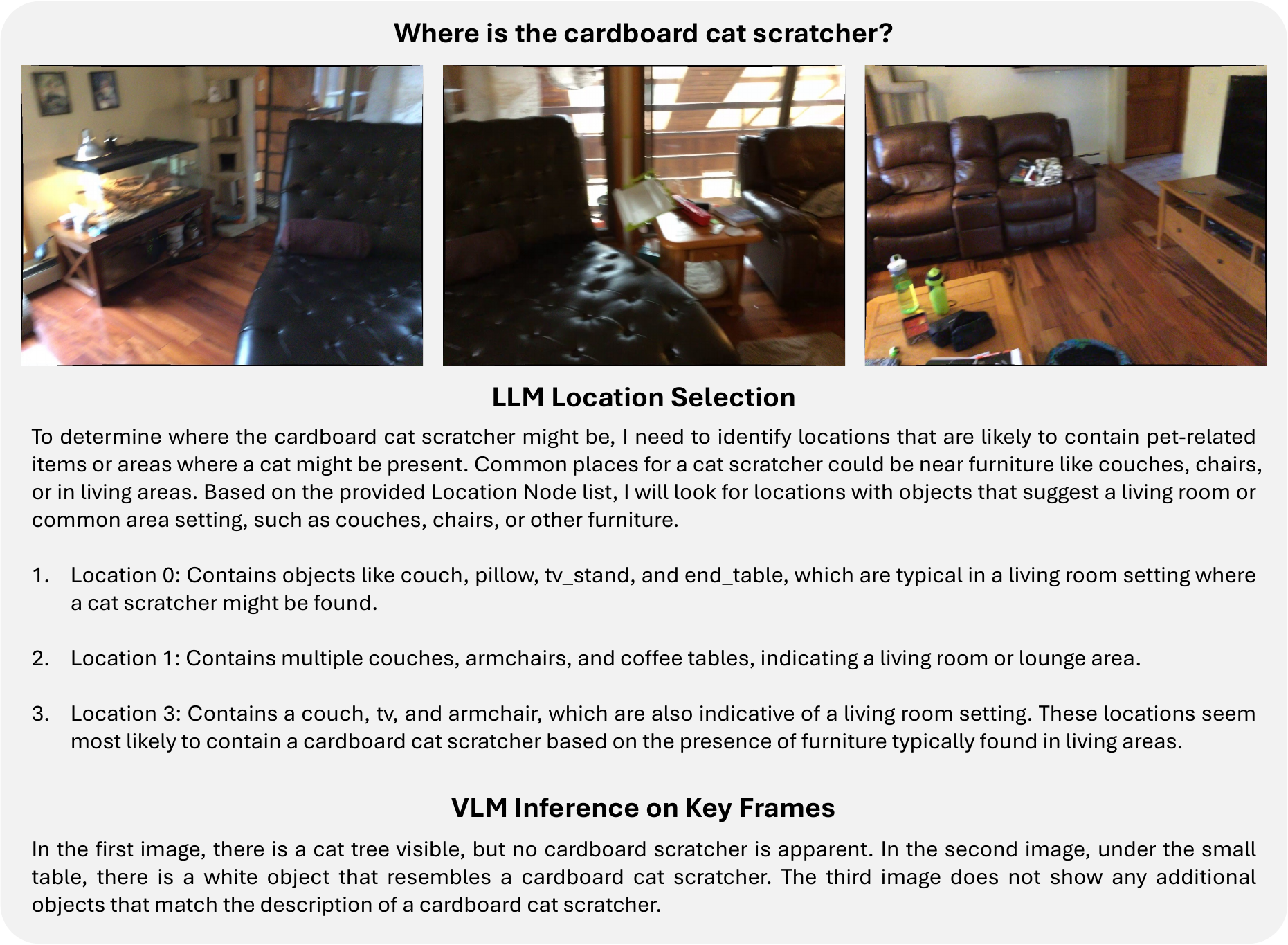}
    \caption{A failure case of an novel object localization question from the OpenEQA. The question asked to locate where is the cardboard cat scratcher, because the cat scratcher is not part of the \texttt{ObjNode} in \model, our LLM conduct its best reasoning effort to select the most possible locations that the cat scratcher can be. However, the final selected frames does not contain a clear view for VLM to answer the question.}
    \label{fig:fail}
\end{figure*}